\documentclass[10pt,twocolumn,a4paper]{esaAI}

\title{A Self-Supervised Task for Fault Detection in Satellite Multivariate Time Series}

\def\authorEmail{carlo.cena@polito.it}

\author[1,2]{Carlo Cena\thanks{Corresponding author. E-Mail: \authorEmail}}
\author[2]{Silvia Bucci}
\author[2]{Alessandro Balossino}
\author[1]{Marcello Chiaberge}
\affil[1]{Politecnico di Torino, Torino (TO), ITA}
\affil[2]{Argotec S.R.L., San Mauro (TO), ITA}

\begin{document}

% Creates the title and author list automatically for you!
\makeCustomtitle

\begin{abstract}
% 200 words max
In the space sector, due to environmental conditions and restricted accessibility, robust fault detection methods are imperative for ensuring mission success and safeguarding valuable assets. This work proposes a novel approach leveraging Physics-Informed Real NVP neural networks, renowned for their ability to model complex and high-dimensional distributions, augmented with a self-supervised task based on sensors' data permutation. It focuses on enhancing fault detection within the satellite multivariate time series. The experiments involve various configurations, including pre-training with self-supervision, multi-task learning, and standalone self-supervised training. Results indicate significant performance improvements across all settings. In particular, employing only the self-supervised loss yields the best overall results, suggesting its efficacy in guiding the network to extract relevant features for fault detection. This study presents a promising direction for improving fault detection in space systems and warrants further exploration in other datasets and applications.
\end{abstract}

\section{Introduction}
Today's world is increasingly reliant on satellite technology for navigation \cite{hofmann2007gnss}, communication \cite{rahmat2014technology}, and scientific studies \cite{martini2021domain, deep_space_1}. It is therefore important to ensure the reliability and longevity of these assets. The challenges presented by the space environment, characterized by extreme temperatures, radiation exposure, and limited, if not absent, maintenance opportunities, underscore the critical importance of fault detection in satellite systems \cite{HASAN2022100806, faults_type}. Traditional approaches often rely on predetermined rules or thresholds, which can be inadequate for capturing the complex and dynamic nature of faults in such demanding conditions, or costly model-based approaches \cite{schein2006rule, 4512019}.

\begin{figure}[t]
    \centering
    \includegraphics[width=1.0\columnwidth]{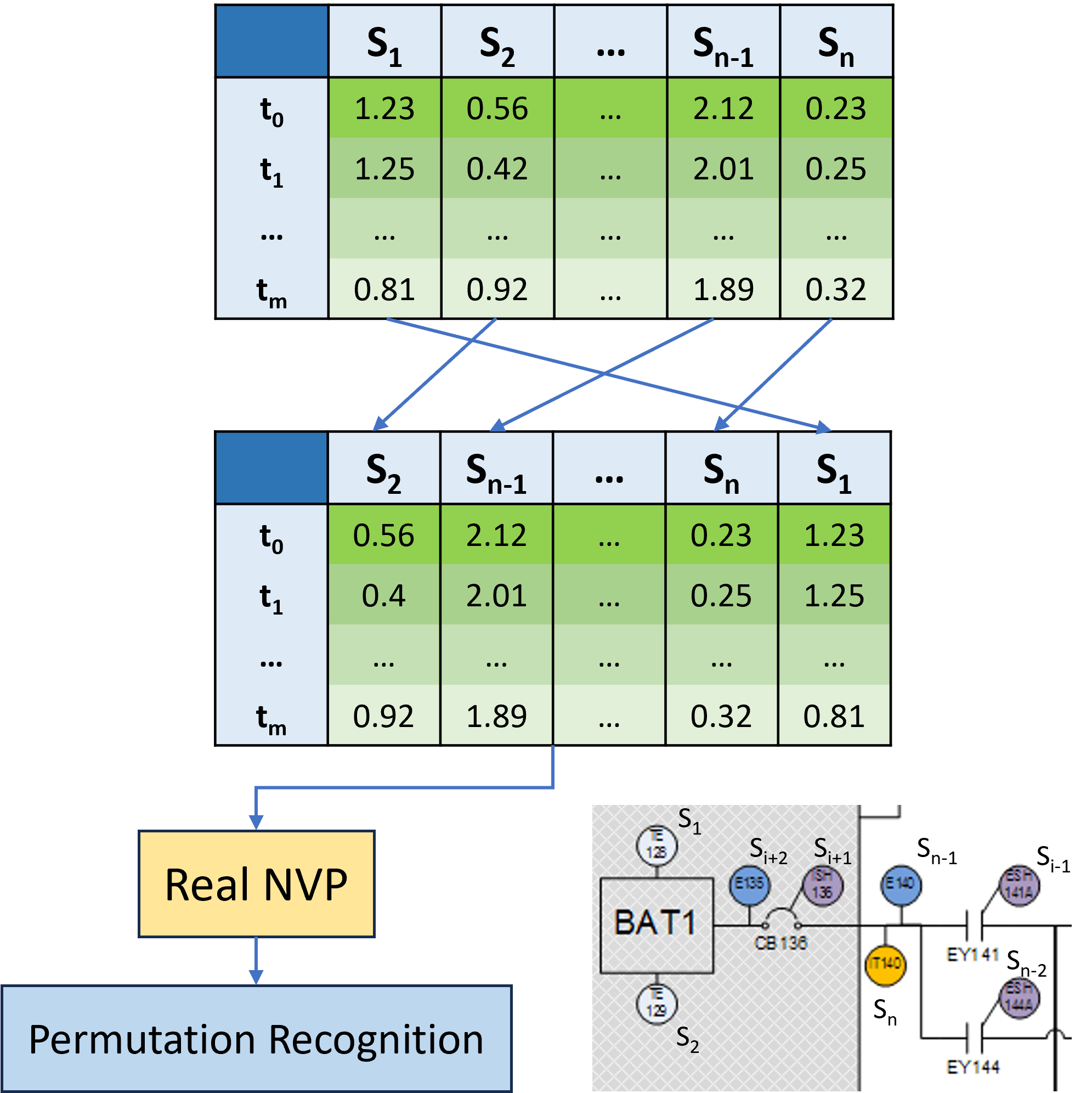}
	\caption{Self-supervised loss: the dataset's columns are permuted and the Real NVP model is trained to predict the correct permutation. As shown in the image patch extracted from the ADAPT circuit, each column is associated with a given sensor of the testbed.}
	\label{fig:selfsup_loss}
\end{figure}
To address these challenges, in recent years we have witnessed a surge of studies in Artificial Intelligence (AI) methodologies for fault detection in space systems, which have shown promising results in automatically identifying and diagnosing anomalies in satellite data. Among these methodologies, the Physics-Informed Real NVP model introduced by \cite{aim2024} demonstrated its ability to model complex distributions while incorporating domain-specific knowledge.

This model integrates principles of physics, specific to the considered dataset, ADAPT \cite{adapt_paper}, into the loss function, enabling the model to capture underlying physical relationships within the data. By leveraging the Normalizing Flow family \cite{norm_flow, 9089305} and affine coupling layers \cite{dinh2014nice}, this model excels in representing high-dimensional distributions, making it well-suited for analyzing multivariate satellite data.

Given the scarcity and cost of labeled data in this sector,
%, to enhance fault detection further it is particularly interesting to reduce the amount of data needed to train new models.
self-supervision offers the advantage of leveraging unlabeled data to guide the learning process, enhancing the model's ability to extract meaningful features relevant to fault detection \cite{zhang2024self}. These approaches typically rely on creating pretext tasks that provide supervisory signals for model training. Their success depends on the effectiveness of the designed pretext tasks, which can be categorized into various paradigms, such as contrastive learning and generative learning \cite{liu2024self}. Contrastive methods, like SimCLR \cite{pmlrv119chen20j}, focus on learning representations by encouraging the model to distinguish between similar and dissimilar data points. Conversely, generative approaches, such as Masked Autoencoders \cite{zhang2022survey}, train models to reconstruct the original data from a corrupted version, forcing them to capture essential features.

Self-supervised tasks should be carefully chosen to exploit the specific structure of the data at hand and steer learning toward robust representations. Examples from other fields include object recognition across different domains using jigsaw puzzles as a self-supervised task \cite{jigsaw_puzzle} or augmentation-aware self-supervised task in the discriminator network of a GAN for robust representation learning \cite{NEURIPS20236464638c}.

Time series data presents unique challenges for self-supervised approaches due to their sequential nature and the importance of capturing temporal dependencies. Traditional methods primarily focus on inter-sample relationships, neglecting the crucial intra-temporal structure within a single time series.

Recent advancements address this gap by incorporating self-supervised tasks to learn inter-sample and intra-temporal relationships \cite{fan2020self, time_permutation}: SelfTime \cite{fan2020self} proposes a framework that explores these relationships simultaneously. It considers separate reasoning heads within the model to analyze both the similarities between different time series and the relationships between time steps within a single series.

The challenges become even more complex when dealing with multivariate time series, where multiple interconnected variables evolve over time. Graph Neural Networks (GNNs) offer a promising solution in this domain \cite{zhang2024self}, as they can capture the relationships between different variables. \cite{saeed2019multi} applies transformations (permutation, scaling, etc.) on sensor data and subsequently tries to differentiate them as a training method for human activity detection.

Recognizing the potential of self-supervised learning techniques to augment existing methodologies, we propose the integration of a self-supervision task into the fault detection process for satellites' Electrical Power System (EPS). We propose to permute the input sensor's measurements (i.e. channels), but differently from \cite{saeed2019multi}, which only aimed at detecting a generic permutation among other transformations, we ask the network to sort them by predicting the permutation's index, similarly to what is done in \cite{jigsaw_puzzle} with image patches.

To the best of our knowledge, this is the first time that the prediction of the permutation applied to the time series features, i.e. sensor's data, is used in the space sector as a self-supervision task, and this is the first study that performs extensive experiments to understand its effect in a multi-task setting and as a standalone loss. To summarize, the contributions of this paper are:
\begin{enumerate}
    \item we introduce a self-supervised task for fault detection in multivariate systems, demonstrating its effectiveness in the space domain;
    \item we evaluate the above-mentioned task in multiple settings, showing its effectiveness both for pre-training and in multi-task training;
    \item we show that when used as a standalone loss this self-supervised task leads to better results on ADAPT \cite{adapt_paper} demonstrating its relevance especially when labels are not available.
\end{enumerate}
Through these contributions, this paper aims to advance the state-of-the-art in satellite fault detection towards more resilient and reliable spacecraft.

\section{Methodology}
Here we delve into the description of the model and the losses used.

The self-supervision loss is computed by permuting the order of the dataset's columns (\(S_{1}, S_{2}, ..., S_{n}\)) (see \cref{fig:selfsup_loss}), and asking the model to predict the permutation applied. 

\begin{figure}[t]
    \centering
    \includegraphics[width=1.0\columnwidth]{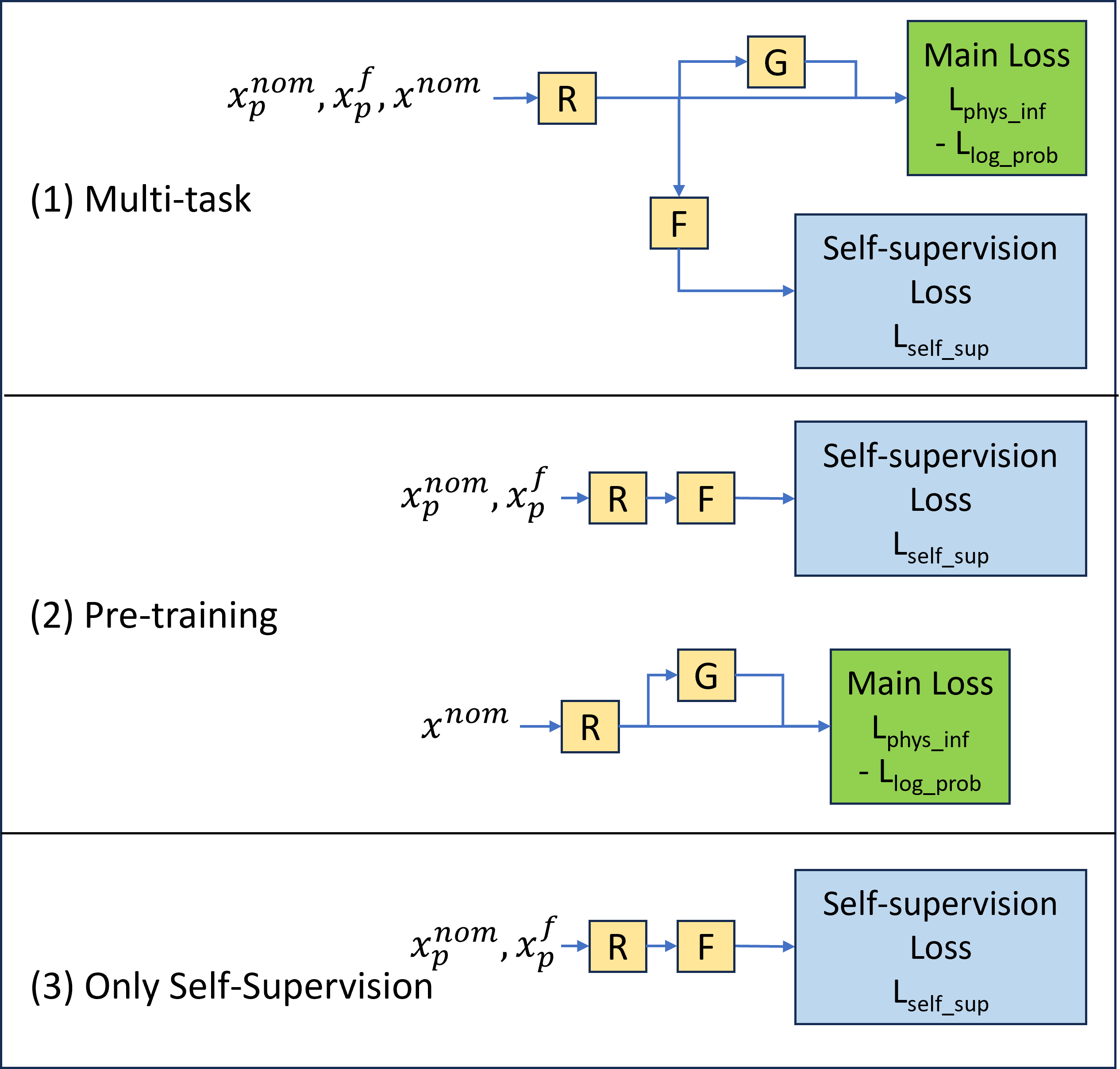}
	%\caption{We considered three different configurations: multi-task, pre-training, and only self-supervision. \(R\), \(F\), and \(G\) represent, respectively, the Real NVP model, a fully connected layer and a Gaussian distribution layer. \(p\) represents the permutation applied to the input and used in the loss function. For the self-supervised task both nominal, \(x_{p}^{nom}\), and fault data, \(x_{p}^f\), were used, while for the main loss only nominal data, \(x^{nom}\), was given as input. The main loss is composed by the default loss for normalizing fault models, i.e. log probability, (\(-L_{\text{log\_prob}}\)), and a physics-informed loss specific to ADAPT, (\(L_{\text{phys\_inf}}\)), while the self-supervised loss, (\(L_{\text{self\_sup}}\)), uses as labels the permutations applied to the input data.}
    \caption{\(R\), \(F\), and \(G\) represent the model, a fully connected and a Gaussian distribution layer. \(p\) is the permutation applied to the input and used in the self-supervised loss, (\(L_{\text{self\_sup}}\)). For the self-supervised task both nominal, \(x_{p}^{nom}\), and fault data, \(x_{p}^f\), were used, while for the main loss only nominal data, \(x^{nom}\), was given as input. The main loss is composed by the default loss for Real NVP, (\(-L_{\text{log\_prob}}\)), and a physics-informed loss specific to ADAPT, (\(L_{\text{phys\_inf}}\)).}
	\label{fig:settings}
\end{figure}

In particular, the loss function is

\begin{equation}
    L_{\text{self\_sup}} = -\frac{1}{N}\sum_{i=1}^N \sum_{p=1}^P y_{i,p} \log\frac{\exp({F(R(x_{i, p}))})}{\sum_{j=1}^P \exp({F(R(x_{i, j}))})} \label{eq:loss_cel}
\end{equation}

where \(N\) is the number of samples, \(P\) is the number of permutations, \(y_{i,p}\) is the label (1 if \(p\) is the correct permutation, 0 otherwise), \(x_{i,p}\) is a vector composed by concatenating 50 rows of the dataset after the application of the given permutation, \(R\) is the Real NVP model and \(F\) a fully-connected layer (see \cref{fig:settings}).

\begin{table*}[htbp]\renewcommand{\arraystretch}{1.0}
    \begin{center}
        \begin{tabular}{c | c | c || c | c | c | c} 
            \hline\hline
            Setting & Configuration & \# Perms & AUROC & FPR95 & F1 & Avg. Prec. \\
            \hline\hline
            Baseline \cite{aim2024} & - & - & 84.92 & 46.81 & 84.51 & 74.64\\
            Multi-task & - & 4000 & 90.43 & 29.36 & 89.42 & 82.72\\
            Multi-task & complete dataset & 8000 & 90.26 & 35.89 & 89.67 & 83.71\\
            Pre-Training & - & 4000 & 90.14 & 30.24 & 89.39 & 82.42\\
            Pre-Training & complete dataset & 8000 & 92.13 & 26.67 & 90.03 & 83.90\\
            Only self-supervision & - & 4000 & 89.96 & 30.20 & 89.34 & 32.40\\
            Only self-supervision & complete dataset & 10000 & 92.50 & 24.13 & 90.41 & 84.68\\
            \hline\hline
        \end{tabular}
        \caption{The best-performing setting is \textit{Only self-supervision} when trained with the complete training set, with both nominal and faulty data.}
        \label{tab:results}
    \end{center}
\end{table*}

We applied our loss to the configuration proposed by \cite{aim2024}: a Real NVP \cite{realnvp_paper} neural network trained with a physics-informed loss, for fault detection.

Real NVP enables the learning of complex, high-dimensional distributions by mapping data to a simpler latent space. Unlike other Normalizing Flow models, Real NVP employs affine coupling layers to capture local dependencies and diverse modes in data, facilitating the generation of realistic samples. Each coupling layer transforms one set of variables while keeping the other unchanged through invertible functions, usually implemented as neural networks. This approach allows for efficient inference and generation of samples.

We tested our self-supervised loss by using it in three different settings, shown in \cref{fig:settings}:
\begin{enumerate}
    \item Multi-task: a multi-task training loop where both losses were used simultaneously;
    \item Pre-Training: a self-supervised pre-training step followed by a fine-tuning step in which \cite{aim2024}'s final loss was used, i.e. our Main Loss;
    \item Only self-supervision: the self-supervision task was used as a standalone loss during training.
\end{enumerate}
In all settings, to compute the self-supervision loss the architecture was augmented with a fully connected layer (\(F\)), with a number of output units equal to the number of permutations (\(P\)) to predict the permutation's index among a given set through a softmax activation function.

\section{Experimental Setup}
In the following, we provide an overview of the architecture used, the dataset considered for the experiments, some implementation details, and the metrics selected to evaluate our solutions.

\textbf{Architecture}: we used the Real NVP with 4 coupling layers \cite{dinh2014nice}, where the neural networks used as translation functions (\textit{t}) and scale functions (\textit{s}) are made up of 2 fully-connected layers with 32 units and a final fully-connected layer with as many units as the dimension of the inputs. As activation, the \textit{t} networks have a linear function, while the \textit{s} neural networks have the tanh function.

\textbf{Dataset}: We used ADAPT \cite{adapt_paper}, a dataset created by NASA Ames Research Center to evaluate EPS fault detection algorithms. We generated 7 splits by randomly splitting the provided files between training and testing data decreasing the number of samples by 66\% with respect to those used in \cite{aim2024} to deal with a more challenging task and to demonstrate the efficacy of our method in conditions where labeled data is scarce and expensive. In all settings we also consider a different configuration (\textit{complete dataset}), in which we used the complete dataset, excluding the test set of the current split, to train the self-supervision task.

\textbf{Implementation Details}: All models have been trained and evaluated on an Intel Xeon Scalable Processors Gold 6130 with 5 different random seeds on all 7 splits using as input a time window of 50 timestamps. Each training was performed keeping 30\% of the training data as validation set and using it for early stopping. Before training, the dataset was scaled between 0 and 1. The self-supervised task requires the permutation of the input features. Given the high number of possible permutations, we decided to create a few sets of permutations of different sizes as follows: for each set, given its target size \textit{P}, we randomly permuted the columns for a number of times well above \textit{P} and then kept the \textit{P} permutations with higher entropy. Specifically, we compute

\begin{equation}
    D = \sum_{i,j=1,i \neq j}^P\sum_{k=1}^n |i_{k} - j_{k}| + \sum_{i=1}^P \sum_{k=1}^n |k - i_{k}| \label{eq:eq_perms}
\end{equation}

where \(n\) is the number of sensors present in the dataset (\cref{fig:selfsup_loss}), \(i\) and \(j\) indicate distinct permutations, \(i_{k}\) (\(j_{k}\)) represents the sensor's index moved in position \(k\) by permutation \(i\) (\(j\)). When selecting the \(P\) permutations we aim to maximize \(D\), as this ensures greater positional deviation from the features' true positions and distinctiveness from other permutations within the set.

\textbf{Metrics}: During inference, the fully connected layer was removed, allowing for the direct use of the log-likelihood values outputted by the Real NVP model's distribution layer (\(G\)) to evaluate the effect of the self-supervised loss. We computed the Area Under the Receiver Operating Curve (AUROC), which measures the model's ability to distinguish between classes by calculating the area under the curve generated by plotting the True Positive Rate against the False Positive Rate with various thresholds; the F1-score, which balances precision and recall by considering both false positives and false negatives; and the False Positive Rate at a 95\% true positive rate (FPR95), which indicates the rate of false alarms when the true positive rate is high.

\section{Results}
In \cref{tab:results} we show, for each setting and configuration, the results obtained evaluating the model on the 7 test sets with 5 different random seeds. 

All the experiments shown in \cref{tab:results} performed better than the baseline, i.e. the neural network and loss proposed by \cite{aim2024} showing that the self-supervised task is useful in driving the network toward the relevant features.

Additionally, the results show that in all settings the \textit{complete dataset} configuration leads to better results in all metrics, but FPR95 in \textit{multi-task}. This is expected as more data is used. More interesting are the results obtained when training using only the self-supervised loss with the complete dataset, which leads to the best results overall. We believe this is because, differently from the final loss in \cite{aim2024}, this loss uses both nominal and faulty data, successfully learning representative features about sensor correlations from both.

Moreover, the experiments showed that the use of a bigger dataset for the self-supervised task always led to an increase in the number of permutations needed for the best results. This may be due to the hardness of the chosen task, which requires a high number of samples to be successfully learned in all cases.

\section{Discussion}
We present a self-supervised task based on feature permutations to pre-train a Physics-Informed - Real NVP neural network for fault detection in multivariate time series. The experiments are performed on several custom splits of the ADAPT dataset.

The results show that using this self-supervised loss leads to improvements with respect to those that can be obtained with previous studies when dealing with a small dataset. This demonstrates the contribution of this work towards more data-efficient AI models, a particularly relevant feature in the space sector, due to the complexities involved in creating a fault detection dataset. Moreover, the proposed task has also been shown to be useful in multi-task settings, as well as a standalone loss.

We believe that the last case is particularly interesting, and we plan on further investigating it to understand the differences between the features extracted by the Real NVP model trained with it and those obtained from the models trained with the log probability and the physics-informed loss. Moreover, we plan on further analyzing the effect of the number of permutations on the self-supervision task in all settings. Finally, the results show that there may be room for additional improvements by using a higher number of samples to pre-train models using bigger sets of possible permutations.

\section*{ACKNOWLEDGMENT}
This work has been developed with the contribution of the Politecnico di Torino Interdepartmental Centre for Service Robotics (PIC4SeR https://pic4ser.polito.it) and Argotec SRL. This publication is part of the project PNRR-NGEU which has received funding from the MUR – DM 117/2023. Computational resources were provided by HPC@POLITO (http://hpc.polito.it).

\printbibliography
\addcontentsline{toc}{section}{References}

\end{document}